\documentclass[10pt,twocolumn,letterpaper]{article}

\usepackage{cvpr}
\usepackage{times}
\usepackage{epsfig}
\usepackage{graphicx}
\usepackage{amsmath}
\usepackage{amssymb}
\usepackage[pagebackref=true,breaklinks=true,letterpaper=true,colorlinks,bookmarks=false]{hyperref}
\usepackage{subfigure}
\usepackage{tabularx}
\usepackage{color}

\makeatletter
\renewcommand{\@thesubfigure}{\hskip\subfiglabelskip}
\makeatother
\usepackage{multirow}

\usepackage[breaklinks=true,bookmarks=false]{hyperref}

\cvprfinalcopy 

\def\cvprPaperID{****} 

\begin{document}

\title{ESIR: End-to-end Scene Text Recognition via Iterative Image Rectification}

\author{Fangneng Zhan\\
Nanyang Technological University\\
50 Nanyang Avenue, Singapore 639798 \\
{\tt\small fnzhan@ntu.edu.sg}
\and
Shijian Lu\\
Nanyang Technological University\\
50 Nanyang Avenue, Singapore 639798 \\
{\tt\small shijian.lu@ntu.edu.sg}
}

\maketitle

\begin{abstract}
Automated recognition of texts in scenes has been a research challenge for years, largely due to the arbitrary variation of text appearances in perspective distortion, text line curvature, text styles and different types of imaging artifacts. The recent deep networks are capable of learning robust representations with respect to imaging artifacts and text style changes, but still face various problems while dealing with scene texts with perspective and curvature distortions. This paper presents an end-to-end trainable scene text recognition system (ESIR) that iteratively removes perspective distortion and text line curvature as driven by better scene text recognition performance. An innovative rectification network is developed which employs a novel line-fitting transformation to estimate the pose of text lines in scenes. In addition, an iterative rectification pipeline is developed where scene text distortions are corrected iteratively towards a fronto-parallel view. The ESIR is also robust to parameter initialization and the training needs only scene text images and word-level annotations as required by most scene text recognition systems. Extensive experiments over a number of public datasets show that the proposed ESIR is capable of rectifying scene text distortions accurately, achieving superior recognition performance for both normal scene text images and those suffering from perspective and curvature distortions.
\end{abstract}

\section{Introduction}
Texts in scenes contain high level semantic information that is very useful in many practical applications such as indoor and outdoor navigation, content-based image retrieval, etc. Accurate and robust recognition of scene texts by machines has been a research challenge for years, largely due to a huge amount of variations in text appearance, the complicated image background, imaging artifacts, etc. The advance of deep learning research and its successes in many computer vision tasks have pushed the boundary of scene text recognition greatly in recent years \cite{mishra2012,Jaderberg15,rodrguez2015,ShiBY17,Lu_pr2017,bai2018,liuyang2018,liu2018mcn,liu2019cse,zhan2018ver,xue2018acc,zhan2019synth,zhan2019sfgan}. On the other hand, the deep learning based approach is still facing various problems while dealing with a large amount of scene texts that suffer from arbitrary perspective distortions and text line curvature.

We design an end-to-end trainable scene text recognition network via iterative rectification (ESIR). The ESIR employs an innovative rectification network that corrects perspective and curvature distortions of scene texts iteratively as illustrated in Fig. 1. The finally rectified scene text image is fed to a recognition network for recognition. The training of the iterative rectification network is driven by better scene text recognition as back-propagated from the recognition network, requiring no other annotations beyond the scene texts as used in most scene text recognition systems.
\begin{figure}[t]
\centering
\includegraphics[scale=0.2]{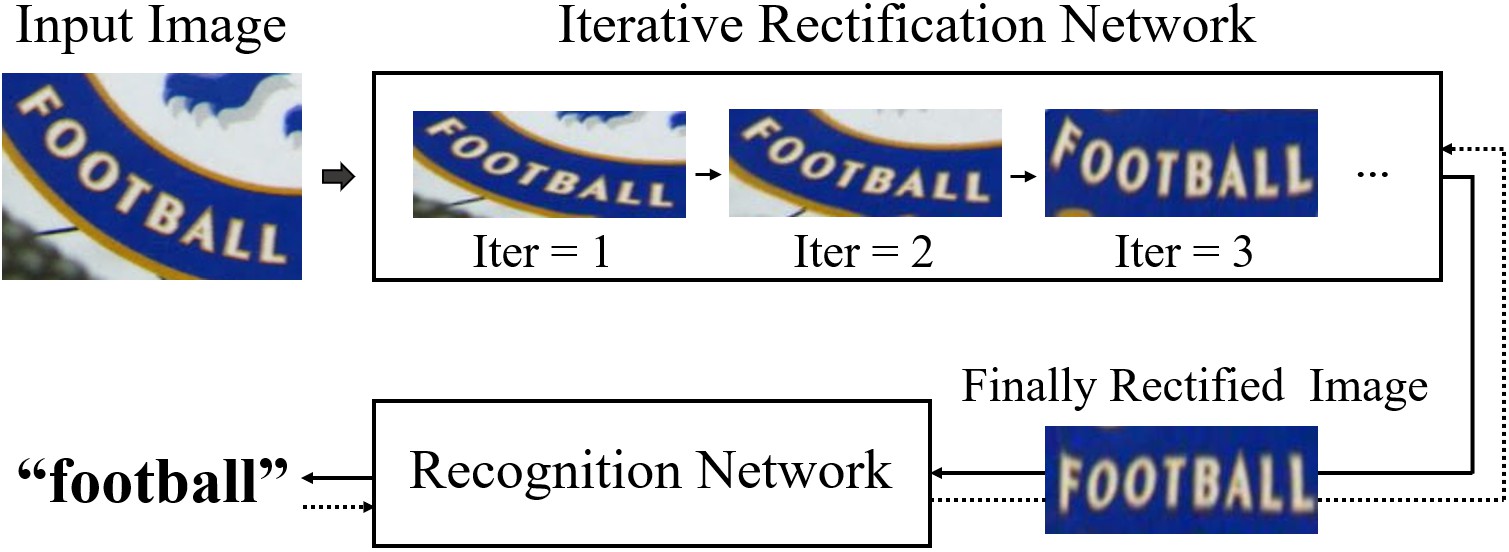}
\vspace{3 pt}
\caption{The proposed system consists of an Iterative Rectification Network that corrects scene text distortions iteratively and a Recognition Network that recognizes the finally rectified scene texts (the images within the Iterative Rectification Network illustrate the iterative scene text rectification process). It is end-to-end trainable as driven by better scene text recognition performance, where the dotted lines show the back propagation of gradients.}
\end{figure}

The proposed ESIR addresses two typical constraints in the scene text recognition problem. The first is \textit{robust} distortion rectification for optimal scene text recognition. To address this challenge, we design a novel line-fitting transformation that is powerful and capable of modeling and correcting various scene text distortions reliably. The line-fitting transformation models the middle line of scene texts by using a polynomial which is able to estimate the pose of either straight or curved text lines flexibly as illustrated in Fig. 2. In addition, it employs line segments which are capable of estimating the orientation and the boundary of text lines in vertical direction reliably. The proposed rectification network is thus capable of correcting not only perspective distortions in straight text lines as in spatial transfer networks \cite{stn} and bag-of-keypoints recognizer \cite{phan2013}, but also various curvatures in curved text lines in scenes. 

The second is \textit{accurate} rectification of perspective and curvature distortions of texts in scenes. To address this challenge, we develop an iterative rectification pipeline that employs multiple feed-forward rectification modules to estimate and correct scene text distortions iteratively. As illustrated in Fig. 2, each iteration takes the image rectified in the last iteration for further distortion estimation as driven by higher scene text recognition accuracy. The iterative rectification is thus capable of producing more accurate distortion correction compared with the state-of-the-art that just performs a single distortion estimation and correction \cite{shi2016,bshi2018aster}. In addition, the iteratively rectified images lead to superior scene text recognition accuracy especially for datasets that contain a large amount of curved and/or perspectively distorted texts, to be described in \textbf{Experiments}.


The contributions of this work are threefold. First, it proposes a novel line-fitting transformation that is flexible and robust for scene text distortion modeling and correction. Second, it designs an iterative rectification framework that clearly improves the scene text rectification and recognition performance with no extra annotations. Third, it develops an end-to-end trainable system that is robust to parameter initialization and achieves superior scene text recognition performance across a number of public datasets. 
\begin{figure}[t]
\centering
\begin{tabular}{ccc}
Input Images &  & Rectified Images \\
\hline
\vspace{-5 pt}
 & & \\
\raisebox{-0.5\height}{\includegraphics[width=0.35\linewidth,height=1.4cm]{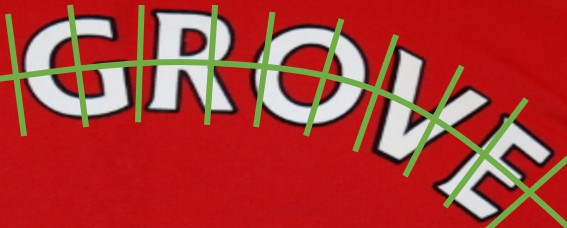}} & $\Rightarrow$ 
& \raisebox{-0.5\height}{\includegraphics[width=0.3\linewidth,height=1cm]{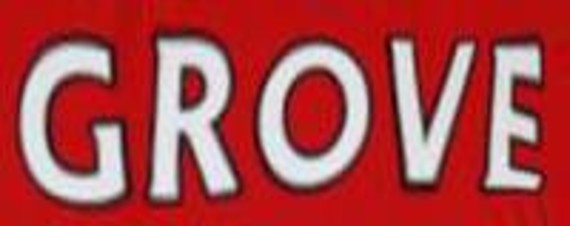}} \\

\vspace{-5 pt}
 & & \\
\raisebox{-0.5\height}{\includegraphics[width=0.35\linewidth,height=1.4cm]{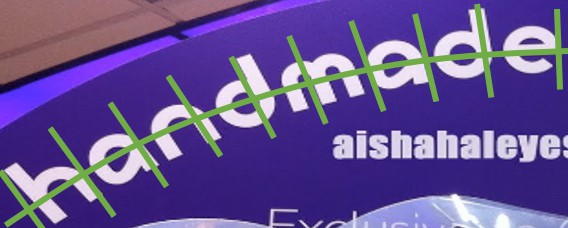}} & $\Rightarrow$ 
& \raisebox{-0.5\height}{\includegraphics[width=0.3\linewidth,height=1cm]{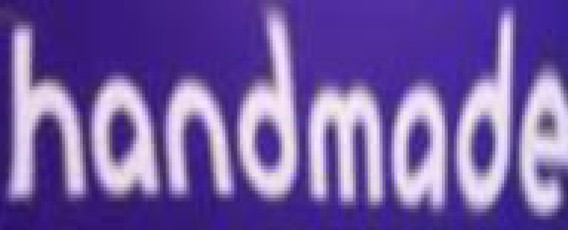}} \\

\vspace{-5 pt}
 & & \\
\raisebox{-0.5\height}{\includegraphics[width=0.35\linewidth,height=1.4cm]{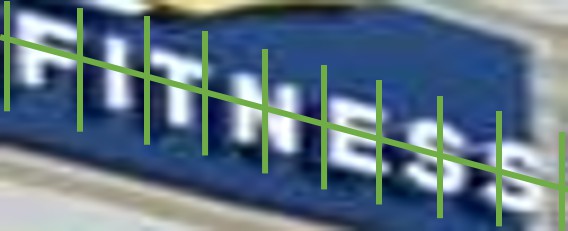}} & $\Rightarrow$ 
& \raisebox{-0.5\height}{\includegraphics[width=0.3\linewidth,height=1cm]{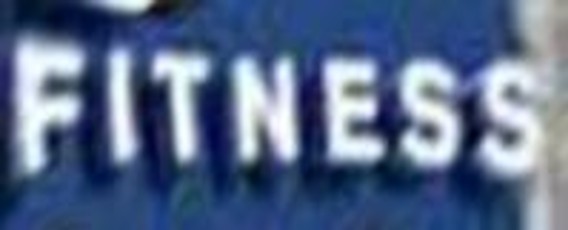}} \\
\end{tabular}
 \vspace{6 pt}
\caption{A novel line-fitting transformation is designed which employs a polynomial to model the middle line of scene texts in horizontal direction, and a set of line segments to estimate the orientation and the boundary of text lines in vertical direction. It helps to model and correct various scene text distortions accurately.}
\end{figure}

\section{Related Work}
\subsection{Scene Text Recognition} Existing scene text recognition work can be broadly grouped into two categories. One category adopts a bottom-up approach that first detects and recognizes individual characters. The other category takes a top-down approach that recognizes words or text lines directly without explicit detection and recognition of individual characters.

Most traditional scene text recognition systems follow a bottom-up approach that first detects and recognizes individual characters by using certain hand-crafted features and then links up the recognized characters into words or text lines using dynamic programming and language models. Different scene character detection and recognition methods have been reported by using sliding window \cite{wang2010,wang2011}, connected components \cite{neumann2012}, extremal regions \cite{neumann2016}, Hough voting \cite{yao2014}, co-occurrence histograms \cite{Lu_pr2016}, etc., but most of them are constrained by the representation capacity of the hand-crafted features. With the advances of deep learning in recent years, various CNN architectures and frameworks have been designed for scene character recognition. For example, \cite{bissacco2013} adopts a fully connected network to recognize characters, \cite{wang2012} uses CNNs for feature extraction, and \cite{Jaderberg15} uses CNNs for unconstrained character recognition. On the other hand, these deep network based methods require localization of individual characters which is resource-hungry and also prone to errors due to complex image background and heavy touching between adjacent characters.

To address the character localization issues, various top-down methods have been proposed which recognize an entire word or text line directly without detecting and recognizing individual characters. One approach is to treat a word as a unique object class and convert the scene text recognition into an image classification problem \cite{Jaderberg16}. In addition, recurrent neural networks (RNNs) have been widely explored which encode a word or text line as a feature sequence and perform recognition without character segmentation. For example, \cite{Lu_accv2014,Lu_pr2017} extract histogram of oriented gradient features across a text sequence and use RNNs to convert them into a feature sequence. \cite{he2016}, \cite{busta2017} and \cite{ShiBY17} propose end-to-end systems that use RNNs for visual feature representation and CTC for sequence prediction. In recent years, visual attention has been incorporated which improves recognition by detecting more discriminative and informative image regions. For example, \cite{lee2016} learns broader contextual information and uses an attention based decoder for sequence generation. \cite{cheng2017} proposes a focus mechanism to eliminate attention drift to improve the scene text recognition performance. \cite{Weilin2018} designs a novel character attention mechanism for end-to-end scene text spotting.

\subsection{Recognition of Distorted Scene Texts} The state-of-the-art combining RNNs and attention has achieved great success while dealing with horizontal or slightly distorted texts in scenes. On the other hand, most existing methods still face various problems while dealing with many scene texts that suffer from either perspective distortions or text line curvatures or both.  
\begin{figure*}
\centering
\includegraphics[width=0.985\linewidth,height=3cm]{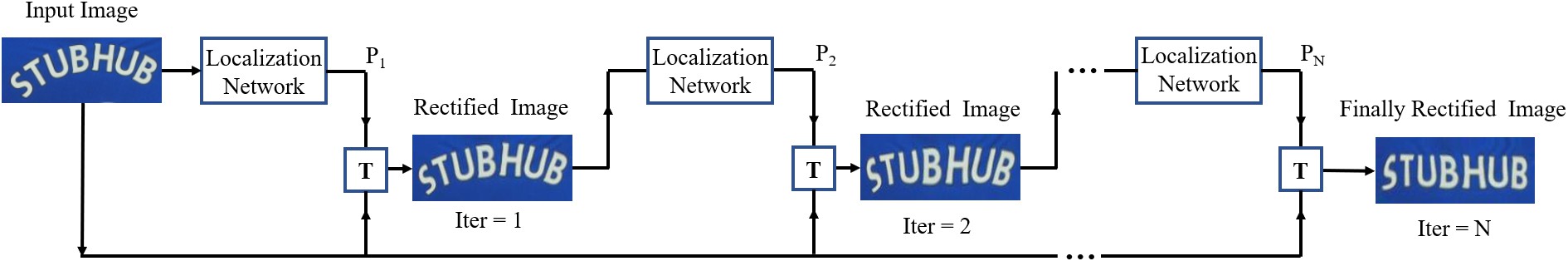}
\caption{The iterative rectification process: \textit{T} denotes a thin plate spline transformation, $P_1$, $P_2$, ... denote transformation parameters that are predicted by the localization network, Iter denotes the number of rectification iterations and \textit{N} is the predefined iteration number.}
\end{figure*}

Prior works dealing with perspective distortions and text line curvatures are limited but this problem has attracted increasing attention in recent years. The very early works \cite{Lu_IVC2005,Paul_2002, lu2006} correct perspective distortions in document texts as captured by digital cameras for better recognition. \cite{phan2013} works with scene texts by using bag of key-points that are tolerant to perspective distortions. These early systems achieve limited successes as they use hand-crafted features and also require character-level information. The recent works \cite{shi2016,bshi2018aster} also take an image rectification approach but explore spatial transformer networks for scene text distortion correction. Similarly, \cite{bartz2018,liu2018} integrate the rectification and recognition into the same network. These recent systems exploit deep convolutional networks for rectification and RNNs for recognition, which require little manually crafted features or extra annotations and have shown very promising recognition performance.

Our proposed technique adopts a rectification approach for robust and accurate recognition of scene texts with perspective and curvature distortions. Different from existing rectification based works~\cite{shi2016,bshi2018aster,bartz2018,liu2018}, it corrects distortions in an iterative manner which helps to improve the rectification and recognition greatly. In addition, we propose a novel line-fitting transformation that is robust and flexible in scene text distortion estimation and correction. 

Note some attempt has been reported in recent years which handles scene text perspectives and curvature distortions by managing deep network features. For example, \cite{yang2017} presents an auxiliary dense detector to encourage visual representation learning. \cite{cheng2018} describes an arbitrary orientation network that extracts scene text features in four directions to deal with scene text distortions. 

\section{The Proposed Method}
This section presents the proposed scene text recognition technique including iterative scene text rectification network, sequence recognition network and detailed description of network training.

\subsection{Iterative Rectification Network}
The proposed iterative rectification network employs a novel line-fitting transformation and an iterative rectification pipeline for optimal estimation and correction of perspective and curvature distortions of texts in scenes.

\subsubsection{Line-Fitting Transformation}
A novel line-fitting transformation is designed to model the pose of scene texts and correct perspective and curvature distortions for better scene text recognition. As illustrated in Fig. 2, the fitting lines consist of a polynomial that is employed to model the middle line of text lines in horizontal direction, and $L$ line segments that are employed to estimate the orientation and the boundary of text lines in vertical direction. Since most text lines in scenes are either along a straight or smoothly curved line, a polynomial of a certain order is sufficient for the estimation of text line poses in the horizontal direction. In our implemented system, a polynomial of order 4 and a set of 20 line segments are employed for scene text pose estimation.

By setting the image center as the origin and normalizing the x-y coordinate of each pixel within scene text images, the middle line of text lines in scenes can be modeled by a polynomial of order $K$ as follows:
$$y = a_{K}*x^K + a_{K-1}*x^{K-1} + \cdots\ + a_{1}*x + a_{0}$$

The $L$ line segments can be modeled by:
$$y = b_{1,l}*x + b_{0,l} \ |\  r_{l}, \quad l = 1,2, \cdots\ ,L$$
where $r_{l}$ denotes the length of line segments on the two sides of the middle line of text lines in scenes which can be approximated as the same. We therefore have $3L$ parameters for estimating the $L$ line segments. By including the middle line polynomial, the number of parameters to be estimated becomes $3L + K + 1$. 
\renewcommand\arraystretch{1.5}

The proposed rectification network iteratively regresses to estimate the fitting-line parameters by employing a localization network together with image convolutions as illustrated in Fig. 3. Table 1 gives detailed structures of the localization network. It should be noted that the training of the localization network does not require any extra annotation of fitting lines but is completely driven by the gradients that are back-propagated from the recognition network. The underlying principle is that higher recognition performance is usually achieved when scene text distortions are better estimated and corrected.

Once the fitting line parameters are estimated, the coordinates of 2 endpoints of each of the $L$ line segments $\left\{t_{j}|j = 1, \cdots\ ,2L\right\}$ can be determined. Scene text distortions can then be corrected by a thin plate spline transformation \cite{tps} that can be determined based on the estimated line segment endpoints and another $2L$ base points $\left\{t'_{j}|j = 1, \cdots\ ,2L\right\}$ which define the appearance of scene texts within the rectified scene text image:
$$
t'_{j} =
\left \{
\begin{array}{l}
[
-0.5+\frac{j-1}{L-1}, 0.5
]^T
, \quad \quad \quad 1 \le j \le L\\

[
-0.5+\frac{l-(L+1)}{L-1}, -0.5
]^T
, \  L+1 \le j \le 2L
\end{array}
\right.
$$

With $P = \left[t_{1}, t_{2}, \cdots\, t_{2L}\right]^T$ and $P' = \left[
t'_{1}, t'_{2}, \cdots\, t'_{2L}\right]^T$, the transformation parameters can be determined by:
$$
C = 
\left[
\begin{array}{ccc}
S & 1_{2L} & P \\
1^{T}_{2L} & 0 & 0 \\
P^{T} & 0 & 0 \\
\end{array}
\right]^{-1}
\cdot
\quad
\left[
\begin{array}{c}
P' \\
0 \\
0 \\
\end{array}
\right]
$$
where \textit{S}=[\textit{U}(\textit{t}-$t_{1}$), \textit{U}(\textit{t}-$t_{2}$), $\cdots$, \textit{U}(\textit{t}-$t_{2L}$)]$^T$ and $U(r) = r^2logr^2$. For every pixel $t'$ within the rectified scene text image, the corresponding pixel $t$ within the distorted scene text image can thus be determined as follows:
$$t = C \cdot t'$$

With the estimated pixels $\left\{t'_{j}|j = 1, \cdots\ ,2L\right\}$, a grid $G=\left\{t_{j}|j = 1, \cdots\ ,2L\right\}$ can be generated within the distorted scene text image for rectification. A sampler is implemented to produce the rectified scene text image by using the determined grid, where the value of the pixel $t'$ is bilinearly interpolated from the pixels near $t$ within the distorted scene text image. The sampler can back propagate the image gradients as it is fully differentiable.

\begin{table}[t]
\caption{The structure of the Rectification Network in Fig. 1}
\centering 
\begin{tabular}{|c|c|c|} 
\hline 
Layers & \multicolumn{1}{c|}{Out Size} & \multicolumn{1}{c|}{Configurations} \\
\hline
Block1 & $16 \times 50$ & $3 \times 3 \ conv, 32, 2 \times 2$ \\
\hline
Block2 & $8 \times 25$ & $3 \times 3 \ conv, 64, 2 \times 2$ \\
\hline
Block3 & $4 \times 13$ & $3 \times 3 \ conv, 128, 2 \times 2$ \\
\hline
FC1 & $512$ & - \\
\hline
FC2 & $\textit{3L+K}$+1 & - \\
\hline
\end{tabular}
\end{table}

\subsubsection{Iterative Rectification}
We develop an iterative rectification pipeline for optimal scene text rectification and recognition. At the first iteration, the rectification network takes the original scene text image as input and rectifies it to certain degrees by using the estimated transformation parameters as described in the last subsection. After that, the rectified scene text image is fed to the same rectification network for further parameter estimation and image rectification. This process repeats until a predefined number of rectification iterations is reached. The finally rectified image is then fed to the sequence recognition network for scene text recognition.

The iterative rectification improves the scene text image rectification performance greatly as to be described in \textbf{Experiments}. On the other hand, it often encounters a critical `boundary effect' problem if the iterative rectification is performed directly without control. In particular, each rectification iteration discards image pixels lying outside of the image sampling region, which accumulates and could lead to the discarding of certain text pixels during the iterative image rectification process. In addition, direct image rectification iteratively often degrades the image clarity severely because of the multiple rounds of bilinear interpolations.

We deal with the `boundary effects' and losing of image clarity by designing a novel network structure as illustrated in Fig. 3. In particular, the rectification network consists of localization networks for estimating rectification parameters, and a thin plate spline transformation $T$ that employs the estimated rectification parameters to generate rectified scene text images. During the iterative image rectification process, the intermediately rectified scene text image is used for parameter estimation only, and the original instead of intermediately rectified scene text image is fed to the transformation module $T$ for rectification consistently. With this new design, `boundary effect' accumulation can be avoided and image clarity will not be degraded, both help to improve the scene text recognition performance greatly as to be presented in \textbf{Experiments}. 

\subsection{Recognition Network}
The recognition network employs a sequence-to-sequence model with an attention mechanism. It consists of an encoder and a decoder. In the encoder, the input is a rectified scene text image that is re-sized to $32\times100$ pixels. A 53-layer residual network \cite{resnet} is used to extract features, where each residual unit consists of a $1\times1$ convolution and a $3\times3$ convolution operations. A $2\times2$ stride convolution is implemented to down-sample feature maps in the first two residual blocks. The convolution stride is then changed to $2\times1$ in all following residual blocks and this helps to reserve more information along the horizontal direction and is also very useful for distinguishing neighboring characters. The residual network is followed by two layers of Bidirectional long short-term memory (BiLSTM) each of which has 256 hidden units. The decoder adopts the LuongAttention mechanism \cite{luong2015} which consists of 2-layer attentional LSTMs with 256 hidden units and 256 attention units. During the inference stage, beam search is employed to decode components that maintain $k$ candidates with the highest accumulative scores.

\subsection{Network Training}
The training of image rectification networks is never a simple task. The major issue is that image rectification networks are sensitive to parameter initialization as frequently encountered and shared in prior studies \cite{shi2016,bshi2018aster}. In particular, random parameter initialization often leads to network convergence problems because it is liable to produce highly distorted scene text images that ruin the training of the recognition network and further the rectification network (note that the training of the rectification network is driven by the scene text recognition performance).

We address the network initialization problem by avoiding direction prediction of $P$ as defined in \textbf{Line-Fitting Transformation}. Instead, we define an auxiliary $P_{0}$ that equals $P'$ at the beginning and make $P = P_{0} + \triangle P$, where $\triangle P$ is predicted by the rectification network iteratively. By assigning a small value to $\triangle P$, the initial $P$ will have a similar value as $P'$. This approach avoids generating highly distorted scene text image at the beginning stage and improves the network convergence greatly. Additionally, the gradual learning of $P$ via iterative estimation of $\triangle P$ (instead of direct prediction of $P$) makes the rectification network training smooth and stable.

\section{Experiments}
\subsection{Datasets and Metrics}
This section describes a list of datasets and evaluation metrics that are used in the experiments.
\renewcommand\arraystretch{1.6} 
\begin{table*}[t]
\caption{Scene text recognition performance over the datasets ICDAR2013, ICDAR2015, IIIT5K, SVT, SVTP and CUTE, where ``50" and ``1K" in the second row denote the lexicon size and ``None" means no lexicon used. The used network backbone and training data are given in [] at the end of each method, where “SK” and “ST” denote the Synth90K and SynthText datasets, respectively.}
\centering 
\begin{tabular}{|l|c|c|ccc|cc|c|c|} 
\hline 
\multirow{2}{*}{Methods} & \multicolumn{1}{c|}{ICDAR2013} & \multicolumn{1}{c|}{ICDAR2015} & \multicolumn{3}{c|}{IIIT5K} & \multicolumn{2}{c|}{SVT} & \multicolumn{1}{c|}{SVTP} & \multicolumn{1}{c|}{CUTE}\\
\cline{2-10}
 & None & None & 50 & 1k & None & 50 & None & None & None \\\hline
Wang \cite{wang2012} \@ [-] & - & - & - & - & - & 70.0 & - & - & - \\
Bissacco \cite{bissacco2013} \@ [-] & 87.6 & - & - & - & - & - & - & - & - \\
Yao \cite{yao2014} \@ [-] & - & - & 80.2 & 69.3 & - & 75.9 & - & - & - \\
AlmazÂ´an \cite{almazan2014} \@ [-] & - & & 91.2 & 82.1 & - & 89.2 & - & - & - \\
Gordo \cite{gordo2015} \@ [-] & - & & 93.3 & 86.6 & - & 91.8 & - & - & - \\
Jaderberg \cite{Jaderberg15} \@ [VGG, SK] & 81.8 & - & 95.5 & 89.6 & - & 93.2 & 71.7 & - & - \\
Jaderberg \cite{Jaderberg16} \@ [VGG, SK] & 90.8 & - & 97.1 & 92.7 & - & 95.4 & 80.7 & - & - \\
Shi \cite{shi2016} \@ [VGG, SK] & 88.6 & - & 96.2 & 93.8 & 81.9 & 95.5 & 81.9 & 71.8 & 59.2 \\
Yang \cite{yang2017} \@ [VGG, Private] & - & - & 97.8 & 96.1 & - & 95.2 & - & 75.8 & 69.3 \\
Cheng \cite{cheng2017} \@ [ResNet, SK+ST] & {\bfseries 93.3} & 70.6 & 99.3 & 97.5 & 87.4 & 97.1 & 85.9 & 71.5 & 63.9 \\
Cheng \cite{cheng2018} \@ [VGG, SK+ST] & - & 68.2 & 99.6 & 98.1 & 87.0 & 96.0 & 82.8 & 73.0 & 76.8 \\
Shi \cite{bshi2018aster} \@ [ResNet, SK+ST] & 91.8 & 76.1 & {\bfseries 99.6} & {\bfseries 98.8} & {\bfseries 93.4} & 97.4 & 89.5 & 78.5 & 79.5 \\

\hline
ESIR \@ [VGG, SK] & 87.4 & 68.4 & 95.8 & 92.9 & 81.3 & 96.7 & 84.5 & 73.8 & 68.4 \\
ESIR \@ [ResNet, SK] & 89.1 & 70.1 & 97.8 & 96.1 & 82.9 & 97.1 & 85.9 & 75.8 & 72.1 \\
ESIR \@ [ResNet, SK+ST] & 91.3 & {\bfseries 76.9} & {\bfseries 99.6} & {\bfseries 98.8} & 93.3 & {\bfseries 97.4} & {\bfseries 90.2} & {\bfseries 79.6} & {\bfseries 83.3} \\\hline
\end{tabular}
\end{table*}

\subsubsection{Datasets}
All ESIR models to be evaluated are trained by using the Synth90K and SynthText, and there is no fine-tuning by using any third dataset. The ESIR models are evaluated over 6 scene text datasets including 3 normal datasets ICDAR2013, IIIT5K and SVT where most scene texts are almost horizontal, and 3 distorted datasets ICDAR2015, SVTP and CUTE80 where a large amount of scene texts suffer from perspective and curvature distortions. The 6 datasets are publicly accessible which have been widely used for evaluations in scene text recognition research.

\textbf{Synth90K} \cite{jaderberg2014} contains 9 million synthetic text images with a lexicon of 90K, and it has been widely used for training scene text recognition models. It has no separation of training and test data and all images are used for training.

\textbf{SynthText} \cite{gupta2016} is the synthetic image dataset that was created for scene text detection research. It has been widely used for scene text recognition research as well by cropping text image patches using the provided annotation boxes. State-of-the-art methods crop different amounts for evaluations, e.g. \cite{cheng2018} crops 4 million, Shi \cite{bshi2018aster} crops over 7 million, etc. We crop 4 million text image patches from this dataset which are at lower end for fair benchmarking.

\textbf{ICDAR2013} \cite{icdar2013} is used in the Robust Reading Competition in the International Conference on Document Analysis and Recognition (ICDAR) 2013. It contains 848 word images for model training and 1095 for testing.

\textbf{ICDAR2015} \cite{icdar2015} was used in the Robust Reading Competition under ICDAR 2015. It contains incidental scene text images that are captured without preparation before capturing. 2077 text image patches are cropped from this dataset, where a large amount of cropped scene texts suffer from perspective and curvature distortions.

\textbf{IIIT5K} \cite{iiit5k} consists of 2000 training images and 3000 test images that are cropped from scene texts and born-digital images. Each word image in this dataset has a 50-word lexicon and a 1000-word lexicon, where each lexicon consists of a ground-truth word and a set of randomly picked words.

\textbf{SVT} \cite{wang2011} is collected from the Google Street View images that were used for scene text detection research. 647 words images are cropped from 249 street view images and words within most cropped word images are almost horizontal. Each word image has a 50-word lexicon.

\textbf{SVTP} \cite{phan2013} consists of 639 word images that are cropped from the SVT images. Most images in this dataset are heavily distorted by perspective distortions which are specifically picked for evaluation of scene text recognition under perspective views. Each word image has a 50-word lexicon as inherited from the SVT dataset.

\textbf{CUTE} \cite{risnumawan2014} consists of 288 word images where most cropped scene texts are curved. All word images are cropped from the CUTE dataset which contains 80 high-resolution scene text images that are originally collected for the scene text detection research. No lexicon is provided for the 288 word images in this dataset.

\subsubsection{Metrics}
We follow the protocol and evaluation metrics that have been widely used in scene text recognition research \cite{cheng2017,bshi2018aster}. In particular, the recognition covers 68 characters including 10 digits, lower-case letters and 32 ASCII punctuation marks. In evaluation, only digits and letters are counted and the rest is directly discarded. If a lexicon is provided, the lexicon word that has the minimum edit distance with the predicted word is selected. In addition, evaluations are based on the correctly recognized words (CRW) which can be determined based on the ground truth transcription.

\subsection{Implementation}
The proposed scene text recognition network is implemented by using the Tensorflow framework. The ADADELTA is adopted as optimizer which employs adaptive learning rate and weighted cross-entropy in sequence loss calculation. The network is trained in 1 million iterations with a batch size of 64. In addition, the network training is performed on a workstation with one Intel Core i7-7700K CPU, one NVIDIA GeForce GTX 1080 Ti graphics card with 12GB memory and 32GB RAM.

Three ESIR models are trained for evaluations and benchmarking with state-of-the-art techniques. The first is a baseline model \textbf{ESIR [VGG, SK]} as shown in Table 2, which uses VGG as the network backbone and the Synth90 as the training data. The second model as denoted by \textbf{ESIR [ResNet, SK]} uses the same training data but ResNet as the network backbone. The third model as denoted by \textbf{ESIR [ResNet, SK+ST]} uses ResNet as the network backbone but a combination of the Synth90K and SynthText as training data, largely for benchmarking with state-of-the-art models such as ASTER and AON that also use a combination of the two datasets in training. All three ESIR models are trained under the same parameters setting: rectification iteration number: 5; number of line segments: 20; order of the middle line polynomial: 4.

\subsection{Experimental Results}
\subsubsection{Rectification and Recognition}
The proposed ESIR has been evaluated extensively over the 6 public datasets as described in \textbf{Dataset} that contain both normal scene text images and scene text images with a variety of perspective and curvature distortions. In addition, it has been benchmarked with a number of state-of-the-art scene text recognition techniques that employ rectification, feature learning techniques, etc. as described in \textbf{Related Work}. Table 2 shows experimental results.
\newcommand{\tabincell}[2]{\begin{tabular}{@{}#1@{}}#2\end{tabular}}
\begin{figure}[t]
\begin{tabular}{ccc}
Input Images & Rectified Images & \tabincell{c}{w/o Rectification \vspace{-10 pt} \\ with Rectification} \\
\hline
\vspace{-20 pt}
 & & \\
\raisebox{-0.5\height}{\includegraphics[width=0.24\linewidth,height=1cm]{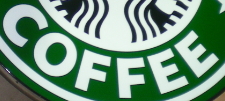}} 
& \raisebox{-0.5\height}{\includegraphics[width=0.24\linewidth,height=0.8cm]{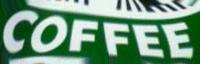}}
& \tabincell{c}{\bfseries \textcolor[rgb]{0,1,0}{offe}\textcolor[rgb]{1,0,0}{x} \vspace{-10 pt} \\ \bfseries \textcolor[rgb]{0,1,0}{coffee}}
\\
\vspace{-20 pt}

 & & \\
\raisebox{-0.5\height}{\includegraphics[width=0.24\linewidth,height=1cm]{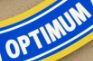}} 
& \raisebox{-0.5\height}{\includegraphics[width=0.24\linewidth,height=0.8cm]{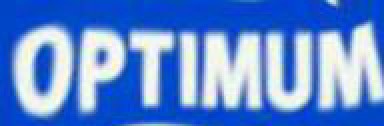}}
& \tabincell{c}{\bfseries \textcolor[rgb]{0,1,0}{opim}\textcolor[rgb]{1,0,0}{inx} \vspace{-10 pt} \\ \bfseries \textcolor[rgb]{0,1,0}{optimum}}
\\
\vspace{-20 pt}

 & & \\
\raisebox{-0.5\height}{\includegraphics[width=0.24\linewidth,height=1cm]{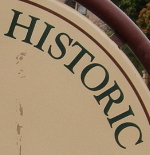}} 
& \raisebox{-0.5\height}{\includegraphics[width=0.24\linewidth,height=0.8cm]{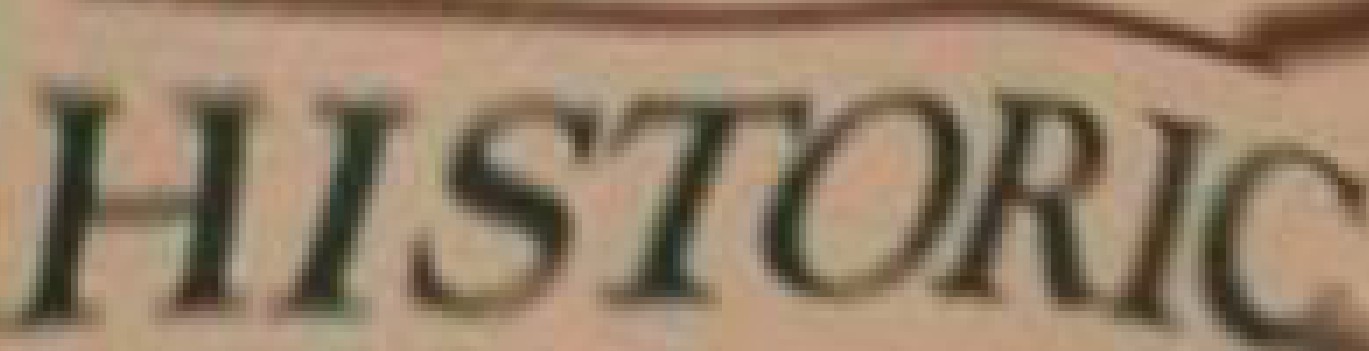}}
& \tabincell{c}{\bfseries \textcolor[rgb]{1,0,0}{mnud}\textcolor[rgb]{0,1,0}{s}\textcolor[rgb]{1,0,0}{e}\textcolor[rgb]{0,1,0}{r} \vspace{-10 pt} \\ \bfseries \textcolor[rgb]{0,1,0}{historic}}
\\
\vspace{-20 pt}

 & & \\
\raisebox{-0.5\height}{\includegraphics[width=0.24\linewidth,height=1cm]{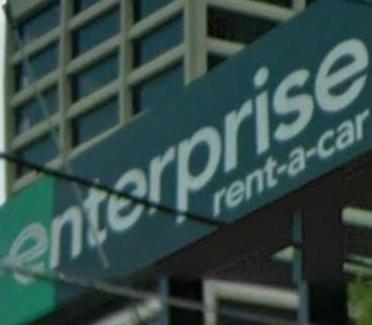}} 
& \raisebox{-0.5\height}{\includegraphics[width=0.24\linewidth,height=0.8cm]{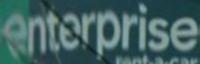}}
& \tabincell{c}{ \bfseries \textcolor[rgb]{1,0,0}{do}\textcolor[rgb]{0,1,0}{s} \vspace{-10 pt} \\ \bfseries \textcolor[rgb]{0,1,0}{enterprise}}
\\
\vspace{-20 pt}

 & & \\
\raisebox{-0.5\height}{\includegraphics[width=0.24\linewidth,height=1cm]{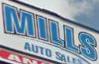}} 
& \raisebox{-0.5\height}{\includegraphics[width=0.24\linewidth,height=0.8cm]{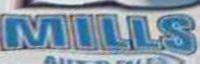}}
& \tabincell{l}{\bfseries \textcolor[rgb]{0,1,0}{m}\textcolor[rgb]{1,0,0}{u}\textcolor[rgb]{0,1,0}{l}\textcolor[rgb]{1,0,0}{t}\textcolor[rgb]{0,1,0}{s} \vspace{-10 pt} \\ \bfseries \textcolor[rgb]{0,1,0}{mills}}
\\
\vspace{-20 pt}

 & & \\
\raisebox{-0.5\height}{\includegraphics[width=0.24\linewidth,height=1cm]{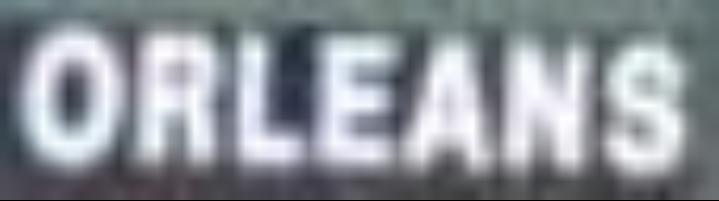}} 
& \raisebox{-0.5\height}{\includegraphics[width=0.24\linewidth,height=0.8cm]{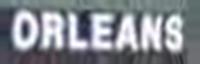}}
& \tabincell{l}{\bfseries \textcolor[rgb]{0,1,0}{orleans} \vspace{-10 pt} \\ \bfseries \textcolor[rgb]{0,1,0}{orleans}}
\\
\vspace{-20 pt}

 & & \\
\raisebox{-0.5\height}{\includegraphics[width=0.24\linewidth,height=1cm]{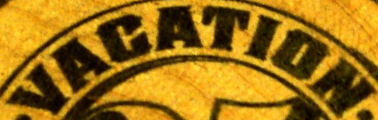}} 
& \raisebox{-0.5\height}{\includegraphics[width=0.24\linewidth,height=0.8cm]{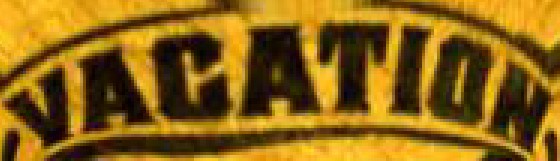}}
& \tabincell{c}{\bfseries \textcolor[rgb]{0,1,0}{a}\textcolor[rgb]{1,0,0}{sg}\textcolor[rgb]{0,1,0}{ati}\textcolor[rgb]{1,0,0}{fy} \vspace{-10 pt} \\ \bfseries \textcolor[rgb]{1,0,0}{t}\textcolor[rgb]{0,1,0}{acation}}
\\
\vspace{-20 pt}

 & & \\
\raisebox{-0.5\height}{\includegraphics[width=0.24\linewidth,height=1cm]{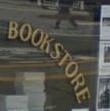}} 
& \raisebox{-0.5\height}{\includegraphics[width=0.24\linewidth,height=0.8cm]{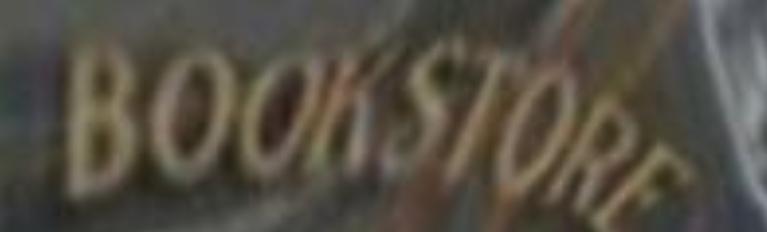}}
& \tabincell{c}{\bfseries \textcolor[rgb]{1,0,0}{srs} \vspace{-10 pt} \\ \bfseries \textcolor[rgb]{0,1,0}{bookstor}\textcolor[rgb]{1,0,0}{f}}
\\
 \vspace{-6 pt}
\end{tabular}
\caption{Illustration of scene text rectification and recognition by the proposed ESIR: For the distorted scene text images from the SVTP and CUTE80 in the first column, the second column shows the finally restored scene text images by the proposed rectification network and the third column shows the recognized texts with and without using the proposed rectification technique, respectively. It is clear that the proposed rectification helps to improve the scene text recognition performance greatly.}
\end{figure}

As Table 2 shows, the \textbf{ESIR [ResNet, SK]} consistently outperforms the \textbf{ESIR [VGG, SK]} across all 6 datasets evaluated due to the use of a more powerful network backbone. In addition, the \textbf{ESIR [ResNET, SK+ST]} consistently outperforms the \textbf{ESIR [VGG, SK]} across the six datasets, demonstrating the value of including more data in network training. By taking a second look at the datasets, we observe that Synth90K mainly consists of English words but few sample images with numbers and punctuation, whereas SynthText contains a large amount of sample images with numbers and punctuation. This could partially explain why the inclusion of the SynthText helps more for the datasets IIT5K and CUTE that contain a large amount of images with numbers and punctuation. 

The proposed ESIR achieves superior scene text recognition performance across the 6 datasets as compared with state-of-the-art techniques. For the three distorted datasets, ESIR outperforms state-of-the-art techniques under all different settings, demonstrating the advantage of the proposed iterative rectification network. In particular, \textbf{ESIR [VGG, SK]} consistently outperforms the \cite{shi2016} over the SVTP and CUTE when similar network backbone and training data are used. \textbf{ESIR [ResNET, SK+ST]} also outperforms \cite{cheng2017} and \cite{bshi2018aster} consistently across the ICDAR2015, SVTP and CUTE under the same setting. For the three normal datasets, ESIR also achieves state-of-the-art performance. In particular, both \textbf{ESIR [VGG, SK]} and \textbf{ESIR [ResNET, SK+ST]} outperform state-of-the-art techniques over the SVT dataset that contains a large amount of low-quality scene texts from street view imagery. For the datasets ICDAR2013, \cite{cheng2017} achieves the best accuracy but it requires character-level bounding box annotations. \cite{Jaderberg16} and \cite{shi2016} outperform the \textbf{ESIR [VGG,SK]} slightly on the ICDAR2013 under the same setting but they only recognize words within a 90K dictionary. \cite{bshi2018aster} achieves the best accuracy on the IIIT5K, but it crops 7.2 million training images from the SynthText whereas we only crop 4 million training images.

Fig. 4 illustrates the scene text rectification and recognition by the proposed ESIR, where the three columns show several sample images from the CUTE and SVTP, the rectified images by using the \textbf{ESIR [VGG, SK]}, and the recognized texts without (at the top) and with (at the bottom) the proposed iterative rectification (incorrectly recognized texts are highlighted in red color), respectively. As Fig. 4 shows, the proposed ESIR is capable of rectifying scene text images with various perspective and curvature distortions in most cases. For the last two severely distorted scene text images, the rectification could be further improved by employing a larger number of rectification iterations (beyond default 5 under the current setting). At the same time, we can see that the proposed ESIR does not degrade scene text images that do not suffer from perspective and curvature distortions as illustrated in the sixth sample image. Further, the proposed ESIR helps to improve the scene text recognition performance greatly as shown in the third column. Note that the recognition here does not use any lexicon, and the recognition performance can be greatly improved by including a lexicon or even a large dictionary, e.g. the mis-recognized 'bookstorf' and 'tacation' from the last two sample images could be corrected if a dictionary is used. 

We conjecture that the ESIR's superior recognition performance especially over the three distorted datasets is largely due to our proposed iterative rectification network. Fig. 5 compares scene text rectifications by our proposed rectification network and two state-of-the-art rectification networks in RARE \cite{shi2016} and ASTER \cite{bshi2018aster}. As Fig. 5 shows, the proposed network produces clearly better rectifications as compared with rectifications by RARE and ASTER. The better rectifications are largely due to the robust line-fitting transformation as well as the iterative rectification framework as described in \textbf{Proposed Method}.
\begin{figure}[t]
\centering
\subfigure {\includegraphics[width=.24\linewidth,height=1cm]{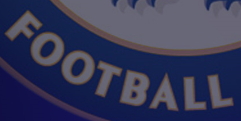}}
\subfigure {\includegraphics[width=.24\linewidth,height=1cm]{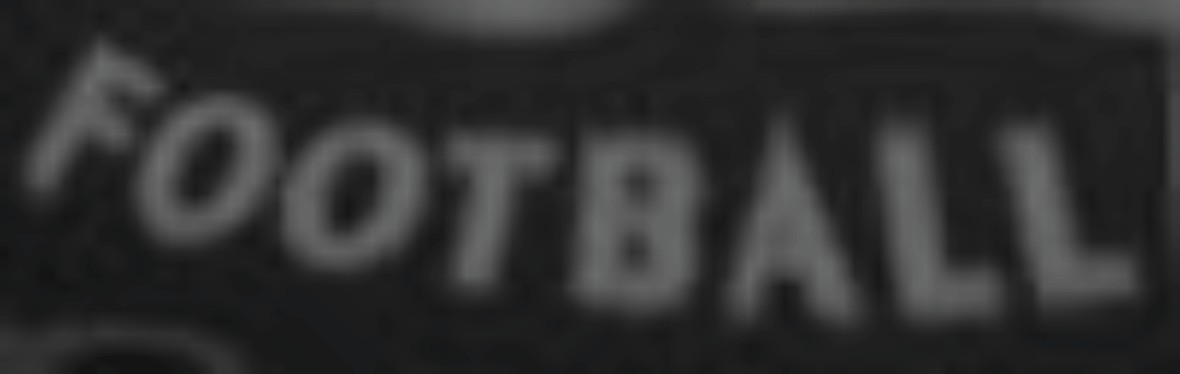}}
\subfigure {\includegraphics[width=.24\linewidth,height=1cm]{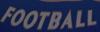}}
\subfigure {\includegraphics[width=.24\linewidth,height=1cm]{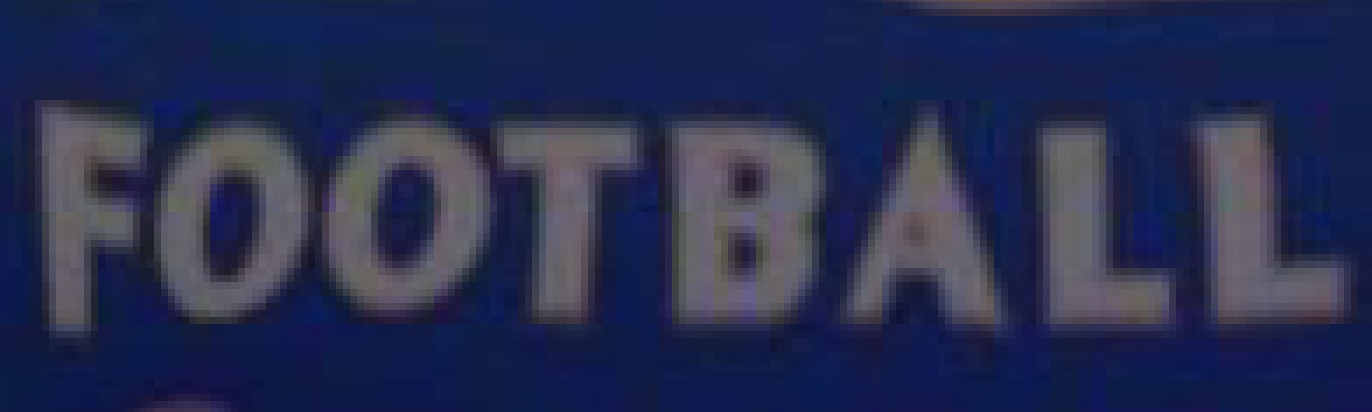}}
\vspace{-8 pt}

\subfigure {\includegraphics[width=.24\linewidth,height=1cm]{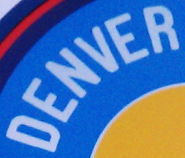}}
\subfigure {\includegraphics[width=.24\linewidth,height=1cm]{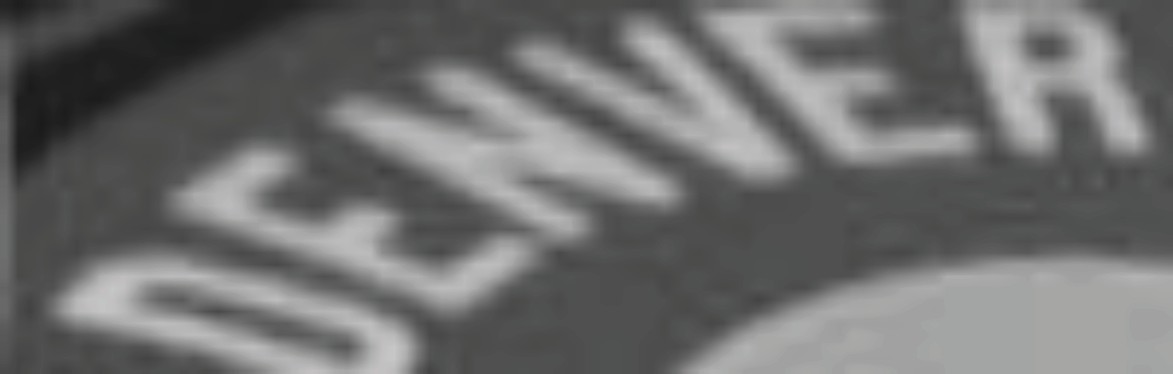}}
\subfigure {\includegraphics[width=.24\linewidth,height=1cm]{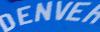}}
\subfigure {\includegraphics[width=.24\linewidth,height=1cm]{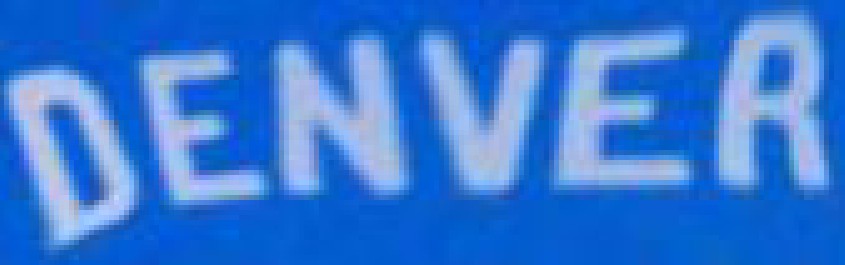}}
\vspace{-8 pt}

\subfigure {\includegraphics[width=.24\linewidth,height=1cm]{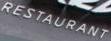}}
\subfigure {\includegraphics[width=.24\linewidth,height=1cm]{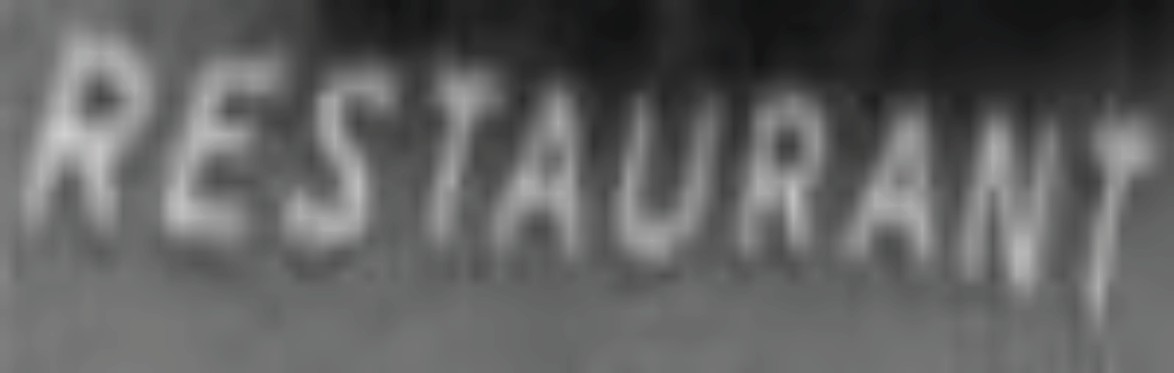}}
\subfigure {\includegraphics[width=.24\linewidth,height=1cm]{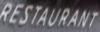}}
\subfigure {\includegraphics[width=.24\linewidth,height=1cm]{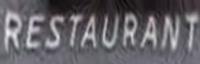}}
\vspace{-8 pt}

\subfigure [Original] {\includegraphics[width=.24\linewidth,height=1cm]{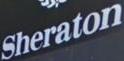}}
\subfigure [By RARE] {\includegraphics[width=.24\linewidth,height=1cm]{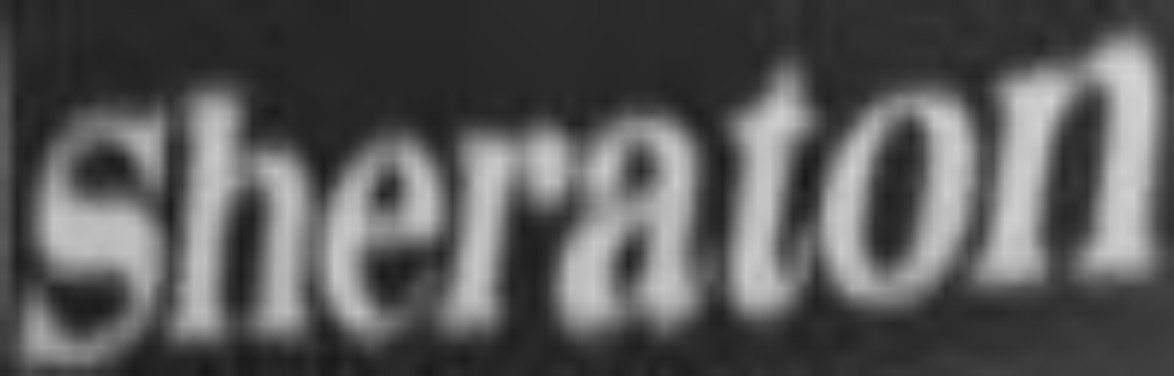}}
\subfigure [By ASTER] {\includegraphics[width=.24\linewidth,height=1cm]{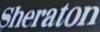}}
\subfigure [By ESIR] {\includegraphics[width=.24\linewidth,height=1cm]{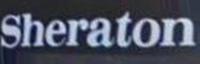}}
 \vspace{3 pt}
\caption{Visual comparison of different scene text rectification methods: For the four sample images in the first column, columns 2-4 show the rectified images by using the RARE, ASTER, and ESIR, respectively. The sample images are from SVTP and CUTE80 which suffer from perspective and curvature distortions as well as complex image background. The proposed ESIR performs clearly better in scene text distortion rectification.}
\end{figure}

\subsubsection{Ablation Analysis}
The performance of the proposed scene text rectification and recognition technique is heavily affected by two key parameters, namely, the number of rectification iterations and the number of line segments as illustrated in Fig. 2. We study these two key parameters separately by using the Synth90K and SynthText as training data consistently. Table 3 shows experimental results, where the first set of experiments fix the number of line segments at 20 but change the rectification iteration number from 1 to 5 and the second set of experiments fix the rectification iteration number at 5 but use 5, 10 and 15 line segments, respectively.

As Table 3 shows, the model performance improves consistently when a larger number of rectification iterations is implemented. In particular, the improvement is more significant at the early stage when the first and second iterations of rectification are implemented i.e. the number of iterations changes from 0 to 1 and from 1 to 2. This can be observed more clearly over the two highly distorted datasets SVTP and CUTE as shown in Table 3. In addition, using a larger number of line segments also helps to improve the scene text recognition performance though the improvement is not as significant as implementing a larger number of rectification iterations. We conjecture that using a larger number of line segments helps to produce better estimations of text line poses which further helps to improve the scene text rectification and recognition performance.
\renewcommand\arraystretch{1.4} 
\begin{table}[t]
\caption{ESIR recognition accuracy under different parameter settings: N denotes the rectification iteration number, L denotes the number of line segments, and Standard Setting uses 5 rectification iterations and 20 line segments. When N or L changes, all other parameters keep the same as the Standard Setting (Synth90K and SynthText are used in training under all settings).}
\centering 
\begin{tabular}{|l|c|c|c|} 
\hline 
Methods & \multicolumn{1}{c|}{ICDAR2015} & \multicolumn{1}{c|}{SVTP} & \multicolumn{1}{c|}{CUTE}\\\hline
\tabincell{l}{Standard Setting: \\ N=5, L=20} & 76.9 & 79.6 & 83.3\\
\hline
\hline
N = 0 & 73.9 & 73.2 & 73.4 \\
N = 1 & 75.8 & 77.3 &  78.8\\
N = 2 & 76.3 & 78.7 &  81.1\\
N = 3 & 76.7 & 79.3 & 82.7 \\
N = 4 & 76.9 & 79.5 & 83.1 \\
\hline
\hline
L = 5 & 75.8 & 78.0 & 81.7 \\
L = 10 & 76.6 & 78.9 & 82.6 \\
L = 15 & 76.9 & 79.3 & 83.0 \\
\hline
\end{tabular}
\end{table}

\subsubsection{Computational Costs}
Though the proposed ESIR performs multiple iterations of rectification, the overall computation costs just increase slightly as compared with state-of-the-art rectification techniques without using iterations \cite{shi2016,bshi2018aster}. In particular, the proposed ESIR with 5 rectification iterations takes 3ms per image in training with a batch-size of 64 and 28ms per image in testing with a batch-size of 1. Under the similar network setting, the ASTER takes 2.4ms per image in training and 20ms per image in testing. The similar computational cost is largely due to the proposed rectification network as shown in Table 1 which is small and computational light as compared with the feature extraction network and the sequence recognition network.

\section{Conclusions}
This paper presents an end-to-end trainable scene text recognition network that is capable of recognizing distorted scene texts via iterative rectification. The proposed network estimates and corrects perspective distortion and text line curvature iteratively as driven by better scene text recognition performance. In particular, a novel line-fitting transformation is designed to estimate the pose of text lines in scenes, and an iterative rectification framework is developed for optimal scene text rectification and recognition. The proposed network is also robust to parameter initialization and does not require extra annotations. Experiments over a number of public datasets demonstrate its superior performance in scene text rectification and recognition. 

The integration with a detection model to realize joint optimization and achieve an end-to-end scene text reading system will be further explored.


{\small
\bibliographystyle{ieee_fullname}
\bibliography{egbib}

\begin{thebibliography}{10}\itemsep=-1pt

\bibitem{almazan2014}
J. Almaz´an, A. Gordo, A. Forn´es, and E. Valveny.
\newblock Word spotting and recognition with embedded attributes.
\newblock {\em TPAMI}, 36(12):2552--2566, 2014.

\bibitem{bai2018}
Fan Bai, Zhanzhan Cheng, Yi Niu, Shiliang Pu, and Shuigeng Zhou.
\newblock Edit probability for scene text recognition.
\newblock In {\em CVPR}, 2018.

\bibitem{bartz2018}
Christian Bartz, Haojin Yang, and Christoph Meinel.
\newblock See: towards semi-supervised end-to-end scene text recognition.
\newblock In {\em AAAI}, 2018.

\bibitem{bissacco2013}
A. Bissacco, M. Cummins, Y. Netzer, and H. Neven.
\newblock Photoocr: Reading text in uncontrolled conditions.
\newblock In {\em ICCV}, 2013.

\bibitem{tps}
Fred~L. Bookstein.
\newblock Principal warps: Thin-plate splines and the decomposition of
  deformations.
\newblock {\em TPAMI}, 11(6), 1989.

\bibitem{busta2017}
Michal Bušta, Lukàš Neumann, and Jirí Matas.
\newblock Deep textspotter: An end-to-end trainable scene text localization and
  recognition framework.
\newblock In {\em ICCV}, pages 2223--2231, 2017.

\bibitem{cheng2017}
Zhanzhan Cheng, Fan Bai, Yunlu Xu, Gang Zheng, Shiliang Pu, and Shuigeng Zhou.
\newblock Focusing attention: Towards accurate text recognition in natural
  images.
\newblock In {\em ICCV}, pages 5076--5084, 2017.

\bibitem{cheng2018}
Zhanzhan Cheng, Yangliu Xu, Fan Bai, Yi Niu, Shiliang Pu, and Shuigeng Zhou.
\newblock Aon: Towards arbitrarily-oriented text recognition.
\newblock In {\em CVPR}, 2018.

\bibitem{Paul_2002}
Paul Clark and Majid Mirmehdi.
\newblock Recognising text in real scenes.
\newblock In {\em IJDAR}, pages 243--257, 2002.

\bibitem{gordo2015}
A. Gordo.
\newblock Supervised mid-level features for word image representation.
\newblock In {\em CVPR}, 2015.

\bibitem{gupta2016}
A. Gupta, A. Vedaldi, and A. Zisserman.
\newblock Synthetic data for text localisation in natural images.
\newblock In {\em CVPR}, 2016.

\bibitem{resnet}
Kaiming He, Xiangyu Zhang, Shaoqing Ren, and Jian Sun.
\newblock Deep residual learning for image recognition.
\newblock In {\em CVPR}, pages 770--778, 2016.

\bibitem{he2016}
Pan He, Weilin Huang, Yu Qiao, Chen~Change Loy, and Xiaoou Tang.
\newblock Reading scene text in deep convolutional sequences.
\newblock In {\em AAAI}, pages 3501--3508, 2016.

\bibitem{Weilin2018}
Tong He, Zhi Tian, Weilin Huang, Chunhua Shen, Yu Qiao, and Changming Sun.
\newblock An end-to-end textspotter with explicit alignment and attention.
\newblock In {\em CVPR}, pages 5020--5029, 2018.

\bibitem{jaderberg2014}
Max Jaderberg, Karen Simonyan, Andrea Vedaldi, and Andrew Zisserman.
\newblock Synthetic data and artificial neural networks for natural scene text
  recognition.
\newblock In {\em NIPS Deep Learning Workshop}, 2014.

\bibitem{Jaderberg15}
Max Jaderberg, Karen Simonyan, Andrea Vedaldi, and Andrew Zisserman.
\newblock Deep structured output learning for unconstrained text recognition.
\newblock In {\em ICLR}, 2015.

\bibitem{Jaderberg16}
Max Jaderberg, Karen Simonyan, Andrea Vedaldi, and Andrew Zisserman.
\newblock Reading text in the wild with convolutional neural networks.
\newblock {\em IJCV}, 2016.

\bibitem{stn}
Max Jaderberg, Karen Simonyan, Andrew Zisserman, and koray kavukcuoglu.
\newblock Spatial transformer networks.
\newblock In {\em NIPS}, 2015.

\bibitem{icdar2015}
Dimosthenis Karatzas, Lluis Gomez-Bigorda, Anguelos Nicolaou, Suman Ghosh,
  Andrew Bagdanov, Masakazu Iwamura, Jiri Matas, Lukas Neumann,
  Vijay~Ramaseshan Chandrasekhar, Shijian Lu, Faisal Shafait, Seiichi Uchida,
  and Ernest Valveny.
\newblock Icdar 2015 competition on robust reading.
\newblock In {\em ICDAR}, pages 1156--1160, 2015.

\bibitem{icdar2013}
D. Karatzas, F. Shafait, S. Uchida, M. Iwamura, S.~R. Mestre, J. Mas, D.~F.
  Mota, J.~A. Almazan, L.~P. de~las Heras, and et al.
\newblock Icdar 2013 robust reading competition.
\newblock In {\em ICDAR}, pages 1484--1493, 2013.

\bibitem{lee2016}
Chen-Yu Lee and Simon Osindero.
\newblock Recursive recurrent nets with attention modeling for ocr in the wild.
\newblock In {\em CVPR}, pages 2231--2239, 2016.

\bibitem{liu2018}
Wei Liu, Chaofeng Chen, and Kwan-Yee~K. Wong.
\newblock Char-net: A character-aware neural network for distorted scene text.
\newblock In {\em AAAI}, 2018.

\bibitem{liuyang2018}
Yang Liu, Zhaowen Wang, Hailin Jin, and Ian Wassell.
\newblock Synthetically supervised feature learning for scene text recognition.
\newblock In {\em ECCV}, 2018.

\bibitem{liu2018mcn}
Zichuan Liu, Guosheng Lin, Sheng Yang, Jiashi Feng, Weisi Lin, and Wang~Ling
  Goh.
\newblock Learning markov clustering networks for scene text detection.
\newblock In {\em CVPR}, 2018.

\bibitem{liu2019cse}
Zichuan Liu, Guosheng Lin, Sheng Yang, Fayao Liu, Weisi Lin, and Wang~Ling Goh.
\newblock Towards robust curve text detection with conditional spatial
  expansion.
\newblock In {\em CVPR}, 2019.

\bibitem{Lu_IVC2005}
Shijian Lu, Ben~Mei Chen, and Chi~Chung Ko.
\newblock Perspective rectification of document images using fuzzy set and
  morphological operations.
\newblock {\em Image and Vision Computing}, pages 541--553, 2005.

\bibitem{lu2006}
Shijian Lu, Ben~M Chen, and Chi~Chung Ko.
\newblock A partition approach for the restoration of camera images of planar
  and curled document.
\newblock {\em Image and Vision Computing}, 24(8):837--848, 2006.

\bibitem{luong2015}
Minh-Thang Luong, Hieu Pham, and Christopher~D. Manning.
\newblock Effective approaches to attention-based neural machine translation.
\newblock In {\em EMNLP}, 2015.

\bibitem{mishra2012}
Anand Mishra, Karteek Alahari, and C.V. Jawahar.
\newblock Scene text recognition using higher order language priors.
\newblock In {\em BMVC}, 2012.

\bibitem{iiit5k}
Anand Mishra, Karteek Alahari, and C.V. Jawahar.
\newblock Scene text recognition using higher order language priors.
\newblock In {\em BMVC}, 2012.

\bibitem{neumann2012}
Lukas Neumann and Jiri Matas.
\newblock Real-time scene text localization and recognition.
\newblock In {\em CVPR}, 2012.

\bibitem{neumann2016}
Lukáš Neumann and Jiří Matas.
\newblock Real-time lexicon-free scene text localization and recognition.
\newblock {\em TPAMI}, pages 1872--1885, 2016.

\bibitem{phan2013}
Trung~Quy Phan, Palaiahnakote Shivakumara, Shangxuan Tian, and Chew~Lim Tan.
\newblock Recognizing text with perspective distortion in natural scenes.
\newblock In {\em ICCV}, 2013.

\bibitem{risnumawan2014}
Anhar Risnumawan, Palaiahankote Shivakumara, Chee~Seng Chan, and Chew~Lim Tan.
\newblock A robust arbitrary text detection system for natural scene images.
\newblock {\em Expert Syst. Appl.}, 41(18):8027--8048, 2014.

\bibitem{rodrguez2015}
J.~A. Rodr´ıguez-Serrano, A. Gordo, and F. Perronnin.
\newblock Label embedding: A frugal baseline for text recognition.
\newblock {\em IJCV}, 2015.

\bibitem{ShiBY17}
B. Shi, X. Bai, and C. Yao.
\newblock An end-to-end trainable neural network for image-based sequence
  recognition and its application to scene text recognition.
\newblock {\em TPAMI}, 39(11):2298--2304, 2017.

\bibitem{shi2016}
Baoguang Shi, Xinggang Wang, Pengyuan Lyu, Cong Yao, and Xiang Bai.
\newblock Robust scene text recognition with automatic rectification.
\newblock In {\em CVPR}, pages 4168--4176, 2016.

\bibitem{bshi2018aster}
Baoguang Shi, Mingkun Yang, Xinggang Wang, Pengyuan Lyu, Cong Yao, and Xiang
  Bai.
\newblock Aster: An attentional scene text recognizer with flexible
  rectification.
\newblock {\em TPAMI}, 2018.

\bibitem{Lu_accv2014}
Bolan. Su and Shijian. Lu.
\newblock Accurate scene text recognition based on recurrent neural network.
\newblock In {\em ACCV}, 2014.

\bibitem{Lu_pr2017}
Bolan. Su and Shijian. Lu.
\newblock Accurate recognition of words in scenes without character
  segmentation using recurrent neural network.
\newblock {\em Pattern Recognition}, pages 397--405, 2017.

\bibitem{Lu_pr2016}
Shangxuan Tian, Ujjwal Bhattacharya, Shijian Lu, Bolan Su, Qingqing Wang,
  Xiaohua Wei, Yue Lu, and Chew~Lim Tan.
\newblock Multilingual scene character recognition with co-occurrence of
  histogram of oriented gradients.
\newblock {\em Pattern Recognition}, pages 125--134, 2016.

\bibitem{wang2011}
K. Wang, B. Babenko, and S. Belongie.
\newblock End-to-end scene text recognition.
\newblock In {\em ICCV}, 2011.

\bibitem{wang2010}
Kai Wang and Serge Belongie.
\newblock Word spotting in the wild.
\newblock In {\em ECCV}, 2010.

\bibitem{wang2012}
Tao Wang, David~J. Wu, Adam Coates, and Andrew~Y. Ng.
\newblock End-to-end text recognition with convolutional neural networks.
\newblock In {\em ICPR}, pages 3304--3308, 2012.

\bibitem{xue2018acc}
Chuhui Xue, Shijian Lu, and Fangneng Zhan.
\newblock Accurate scene text detection through border semantics awareness and
  bootstrapping.
\newblock In {\em ECCV}, pages 355--372, 2018.

\bibitem{yang2017}
Xiao Yang, Dafang He, Zihan Zhou, Daniel Kifer, and C.~Lee Giles.
\newblock Learning to read irregular text with attention mechanisms.
\newblock In {\em IJCAI}, pages 3280--3286, 2017.

\bibitem{yao2014}
Cong Yao, Xiang Bai, Baoguang Shi, and Wenyu Liu.
\newblock Strokelets: A learned multi-scale representation for scene text
  recognition.
\newblock In {\em CVPR}, 2014.

\bibitem{zhan2018ver}
Fangneng Zhan, Shijian Lu, and Chuhui Xue.
\newblock Verisimilar image synthesis for accurate detection and recognition of
  texts in scenes.
\newblock In {\em ECCV}, pages 249--266, 2018.

\bibitem{zhan2019synth}
Fangneng Zhan, Hongyuan Zhu, and Shijian Lu.
\newblock Scene text synthesis for efficient and effective deep network
  training.
\newblock {\em arXiv:1901.09193}, 2019.

\bibitem{zhan2019sfgan}
Fangneng Zhan, Hongyuan Zhu, and Shijian Lu.
\newblock Spatial fusion gan for image synthesis.
\newblock In {\em CVPR}, 2019.

\end{thebibliography}
}

\end{document}